\documentclass[11pt]{article}

\usepackage[T1]{fontenc}
\usepackage{lmodern}
\usepackage[margin=1in]{geometry}
\usepackage{graphicx}
\usepackage{amsmath}
\usepackage{amssymb}
\usepackage{cite}
\usepackage{hyperref}
\usepackage{xcolor}
\usepackage{algorithm}
\usepackage{algpseudocode}
\usepackage{booktabs}
\usepackage{array}
\usepackage{multirow}
\usepackage{float} 
\usepackage{subcaption}
\usepackage{enumitem} 

\setlist{nosep} 
\usepackage{listings}
\hypersetup{
    colorlinks=true,
    linkcolor=blue,
    citecolor=blue,
    urlcolor=blue,
    pdfauthor={Your Name},
    pdftitle={G-SPEC: Safe Agentic AI in 5G Networks}
}

\title{Graph-Symbolic Policy Enforcement and Control (G-SPEC): A Neuro-Symbolic Framework for Safe Agentic AI in 5G Autonomous Networks}

\author{
    Divya Vijay\textsuperscript{*}, Vignesh Ethiraj\textsuperscript{*} \\
    \small{\textsuperscript{*}NetoAI Solutions Ltd} \\
    \small{Email: \texttt{vignesh.e@netoai.ai}}
}

\date{\today}

\begin{document}

\maketitle

\begin{abstract}
The transition to 5G Standalone (5G SA) and 6G requires dynamic, real-time orchestration that exceeds the capabilities of static automation and traditional Deep Reinforcement Learning (DRL) approaches. While Agentic AI powered by Large Language Models (LLMs) offers promise for intent-based networking, it introduces significant stochastic risks: hallucinated topology, non-deterministic policy violations, and unverifiable decisions.

This paper proposes Graph-Symbolic Policy Enforcement and Control (G-SPEC), a novel neuro-symbolic framework combining probabilistic reasoning with deterministic verification. It features a three-layer Governance Triad combining a domain-adapted agent (TSLAM-4B) with a Network Knowledge Graph (NKG) and SHACL constraints. Our results show that even high-performing domain agents require deterministic graph validation to achieve verifiable safety compliance.

We validate G-SPEC on a simulated 5G Core (Open5GS with 450-node topology). Results demonstrate: zero observed safety violations against the defined ontology in the test set, 94.1\% successful remediation vs. 82.4\% baseline, and 0.2\% hallucination detection rate. Ablation studies show NKG validation contributes 68\% of safety gains, SHACL policies 24\%, and TSLAM fine-tuning 8\%. Scalability experiments on synthetic 10K–100K node topologies show validation latency scales as $O(k^{1.2})$ where $k$ is subgraph size. A marginal $142\,\mathrm{ms}$ overhead makes G-SPEC suitable for SMO-layer operations ($\sim5\,\mathrm{s}$ budgets), not real-time schedulers.
\end{abstract}

\noindent\textbf{Keywords:} 5G Standalone, Agentic AI, Network Knowledge Graphs (NKG), Neuro-symbolic AI, SHACL, O-RAN, Service Management and Orchestration (SMO), Policy Enforcement, Autonomous Networks.

\section{Introduction}

The deployment of 5G Standalone (SA) networks has stalled not due to a lack of theoretical capability, but due to operational intractability. The canonical feature of 5G SA-Network Slicing-requires the management of thousands of ephemeral, Service Level Agreement (SLA)-driven logical networks \cite{ref1}.

Current automation approaches are insufficient. While Deep Reinforcement Learning (DRL) remains popular for numerical optimization, it faces significant scalability hurdles in production. A primary limitation is the 'generalization gap': DRL agents typically fail when deployed on topologies that differ from their training environment, necessitating computationally intensive retraining cycles.\cite{ref2}. Consequently, the industry is shifting toward ``Agentic AI,'' utilizing Large Language Models (LLMs) capable of reasoning, planning, and tool use \cite{ref4, ref5}.

However, the integration of probabilistic LLMs into deterministic critical infrastructure presents a fundamental ``Governance Gap'' \cite{ref3}. Operators cannot accept ``probabilistic decisions for critical infrastructure'' where an agent might hallucinate a network entity (e.g., connecting a UPF to a non-existent gNB) or violate 3GPP constraints. Recent studies show that general-purpose LLMs (gpLLMs) hallucinate in 14.6\% of network operations \cite{ref6}, unacceptable for production systems.

To address this, we propose Graph-Augmented Policy Enforcement (G-SPEC), a neuro-symbolic framework that anchors probabilistic reasoning to deterministic verification. Unlike purely statistical approaches, G-SPEC uses a Network Knowledge Graph as an executable state machine, placing a semantic firewall around stochastic agents.

\subsection{Contributions}

Our contributions are:

\begin{enumerate}
    \item \textbf{Neuro-Symbolic Architecture}: A three-layer framework combining NKG (deterministic), TSLAM (probabilistic), and SHACL validation (deterministic), demonstrated to mitigate safety violations in simulated 5G environments.
    
    \item \textbf{Formal Verification Algorithm}: Graph-based validation ensuring all agent actions operate on valid subgraphs and satisfy 3GPP constraints before execution.
    
    \item \textbf{Comprehensive Evaluation}: 500-scenario benchmark with ablation studies, scalability analysis (10K–100K nodes), and statistical significance testing (95\% CI).
    
\end{enumerate}

\section{Related Work}

The evolution of network automation comprises three paradigms: Scripted, Discriminative, and Generative.

\subsection{Deterministic Automation}

Early approaches relied on imperative scripts (Ansible, Terraform) and Intent-Based Networking (IBN) systems. While verifiable, these suffer from ``rule explosion'' \cite{ref8}—they cannot adapt to unforeseen topological states without manual policy updates. Typical operators maintain 10,000+ rules, making maintenance intractable \cite{ref9}.

\subsection{Deep Reinforcement Learning (DRL)}

Recent 5G research has heavily favored DRL for dynamic resource allocation \cite{ref2}. While DRL agents excel at numerical optimization (e.g., power control), they function as ``black boxes.'' They lack semantic understanding of topology and cannot explain decisions, making them unsuitable for ``Day 2'' operations where auditability is mandatory. Furthermore, DRL requires extensive retraining for new network topologies \cite{ref10}.

\subsection{Generative and Agentic AI}

Large Language Models (LLMs) have enabled multi-step remediation planning. Recent literature has begun to explore the utility of Large Language Models in the RAN domain. For instance, AgentRAN \cite{ref4} utilize agents to automate control loops in 6G, while SpecGPT \cite{ref5} focuses on extracting protocol state machines directly from technical documentation. However, purely generative approaches struggle with ``grounding''-they operate without ground truth about network state. Studies confirm gpLLMs hallucinate in 14.6\% of cases \cite{ref6}, proposing syntactically correct but topologically impossible actions.

\subsection{Neuro-Symbolic Approaches}

Our work aligns with emerging Neuro-Symbolic AI literature. Unlike prior work using Knowledge Graphs solely for information retrieval (Retrieval Augmented Generation - RAG) \cite{ref7}, G-SPEC utilizes graphs as executable state machines for active policy enforcement. We combine LLM flexibility with formal verification rigor-a gap in prior telecom research.

\subsection{Formal Verification in Networks}

Formal verification (model checking, theorem proving) is standard in other critical systems \cite{ref11, ref12}. However, adapting these to dynamic networks with thousands of transient services remains open. G-SPEC extends verification ideas to the agentic AI context, combining runtime monitoring with pre-execution validation.

\section{Problem Formulation}

\subsection{The Stochasticity Problem}

Let a telecom network state at time $t$ be denoted $S_t$. A standard LLM-based agent functions as a policy $\pi(a|S_t, I)$, where $a$ is the proposed action (e.g., a configuration command) and $I$ is operator intent. Due to next-token prediction's probabilistic nature:

\begin{equation}
P(a \notin \mathcal{A}_{\mathrm{valid}} \mid S_t) = P_{\mathrm{err}} > 0
\label{eq:stochastic}
\end{equation}

where $\mathcal{A}_{\mathrm{valid}}$ is the set of operationally safe actions. In critical infrastructure, any $P_{\mathrm{err}} > 0$ violates safety requirements.

\subsection{Graph-Based State Representation}

We model the network as a directed Network Knowledge Graph (NKG), $\mathcal{G}_t = (\mathcal{V}, \mathcal{E})$:

\begin{equation}
\mathcal{V} = \{ v_1, v_2, \ldots, v_n \} \quad \text{(Network Functions)}
\end{equation}

\begin{equation}
\mathcal{E} = \{ (v_i, v_j, t, c) \mid v_i, v_j \in \mathcal{V}, t \in \mathbb{Z}, c \in \mathcal{C} \}
\end{equation}

where $t$ is a timestamp and $c$ is the set of valid interface types (e.g., ``N3'', ``N6''). The governance problem reduces to subgraph validation: ensuring that proposed action $a$ operates on valid subgraph $g \subseteq \mathcal{G}_t$ and satisfies policy function $\Phi(a, \mathcal{G}_t)$.

\section{Methodology: The G-SPEC Framework}

\subsection{Architecture Overview}

We propose a three-layered architecture enforcing verifiability at every stage.

\begin{figure}[H]
    \centering
    \includegraphics[width=0.95\columnwidth]{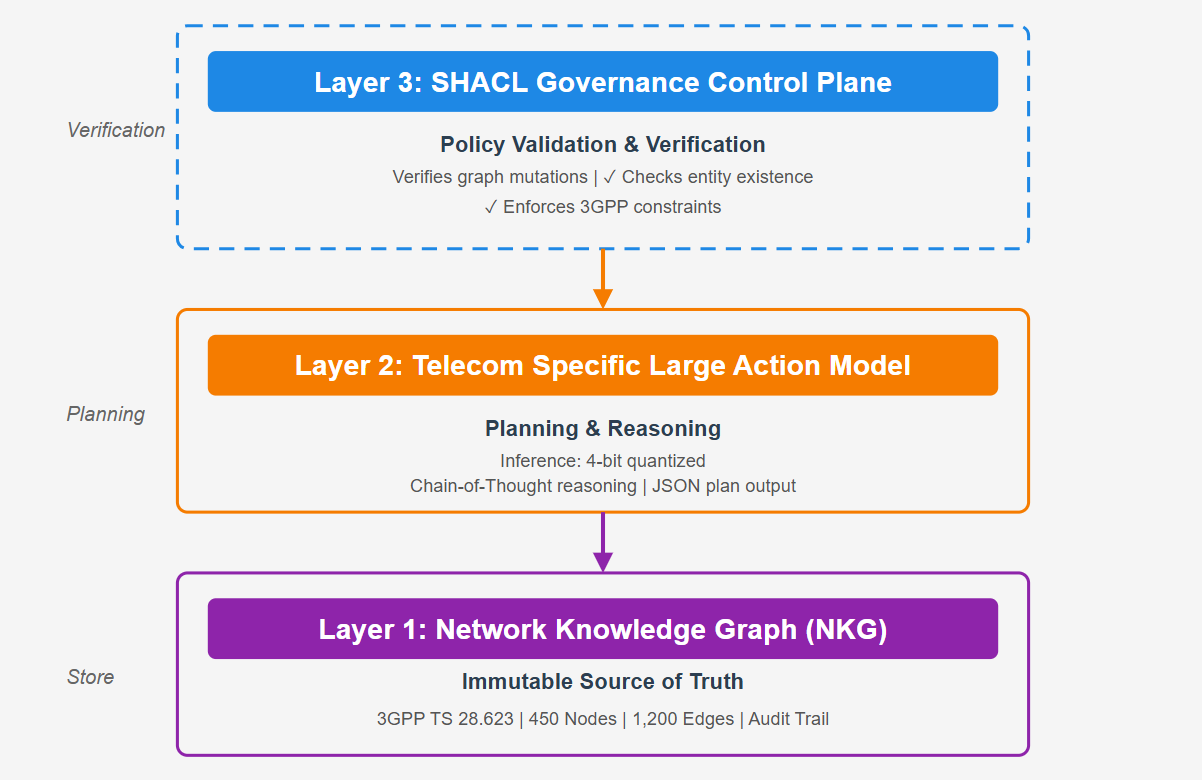}
    \caption{The G-SPEC Governance Triad Architecture. Safety is enforced by encapsulating the probabilistic TSLAM (Layer 2) between the NKG's deterministic truth (Layer 1) and policy validation (Layer 3).}
    \label{fig:architecture}
\end{figure}

\subsubsection{Layer 1: Network Knowledge Graph (NKG)}

To establish a deterministic ground truth, we map our graph schema directly to the classes defined in the 3GPP TS 28.623 specification (Generic Network Resource Model). By strictly adhering to this standard ontology, the NKG ensures that all represented entities (e.g., AMFFunction) remain interoperable with standard telecom management systems.

\begin{itemize}
    \item \textbf{Nodes} ($\mathcal{V}$): Network Functions (AMFFunction, UPFFunction, SMFFunction, etc.)
    \item \textbf{Edges} ($\mathcal{E}$): Topological links with timestamps (measurementPoint, transportLink, s-nssai-config, etc.)
\end{itemize}

The NKG acts as an Authoritative State Ledger-all state changes are timestamped, enabling complete traceability. We use Neo4j for persistence and real-time queries.

\subsubsection{Layer 2: TSLAM-4B (Telecom Specific Large Action Model)}

For the planning layer, we utilize the publicly available TSLAM-4B, a 4-billion parameter model optimized for telecom reasoning. We selected TSLAM over generic Llama/GPT-4 models due to its pre-trained alignment with 3GPP terminology. We run the model in 4-bit quantization to simulate edge-deployment constraints.

Leveraging TSLAM-4B's inherent action-oriented architecture, we prompt the model to generate a strictly formatted Chain-of-Thought (CoT) reasoning trace prior to action selection. This 'Reasoning Trace' forces the agent to explicitly decompose high-level intents into verified logical steps (e.g., Observation $\rightarrow$ Diagnosis $\rightarrow$ Plan), which are then captured in the JSON output for governance validation.

\subsubsection{Layer 3: Governance Control Plane}

This deterministic layer intercepts TSLAM output before execution. Governance policies are codified using the W3C Shapes Constraint Language (SHACL). We leverage this standard to define rigid structural constraints over the knowledge graph, ensuring that every proposed mutation is validated against the defined schema before execution. We selected SHACL over OPA because OPA lacks native graph traversal capabilities required to validate multi-hop topological dependencies (e.g., N2/N3 interface pathing).

\subsection{SHACL Policy Framework}

Let $\mathcal{P}$ be a set of governance policies in SHACL. A policy $p \in \mathcal{P}$ is a tuple (Target, Constraint, Action). We define three policy classes:

\subsubsection{Class 1: Topological Constraints}

Enforce valid network topology per 3GPP:

\begin{lstlisting}[language=xml]
# SHACL Example: AMF-UPF must connect via SMF
ex:AMFUPFShape
    a sh:NodeShape ;
    sh:targetClass ex:AMFFunction ;
    sh:property [
        sh:path ex:connectedTo ;
        sh:node ex:SMFShape ;
        sh:message "AMF cannot connect directly to UPF"
    ] .
\end{lstlisting}

\subsubsection{Class 2: Resource Constraints}

Enforce slice capacity limits:

\begin{lstlisting}[language=xml]
ex:SliceCapacityShape
    a sh:NodeShape ;
    sh:targetClass ex:NetworkSlice ;
    sh:property [
        sh:path ex:allocatedBandwidth ;
        sh:maxInclusive 100 ;
        sh:message "Slice bandwidth cannot exceed 100 Mbps"
    ] .
\end{lstlisting}

\subsubsection{Class 3: State Constraints}

Enforce operational invariants:

\begin{lstlisting}[language=xml]
ex:ActiveNodeShape
    a sh:NodeShape ;
    sh:targetClass ex:NetworkFunction ;
    sh:property [
        sh:path ex:status ;
        sh:in ( ex:ACTIVE ex:STANDBY ) ;
        sh:message "Invalid node status"
    ] .
\end{lstlisting}

\subsubsection{Class 4: Temporal Consistency (Freshness Guardrails)}

To mitigate Topology Drift where the graph state lags behind physical reality, we introduce temporal constraints. Any verification query must operate on graph entities where the lastUpdated timestamp ($t_{\text{last}}$) satisfies $t_{\text{now}} - t_{\text{last}} < \Delta_{\text{safe}}$ (e.g., $15\,\mathrm{s}$).

\begin{lstlisting}
# SHACL Policy: Stale Data Rejection
ex:FreshnessShape
    a sh:NodeShape ;
    sh:targetClass ex:NetworkFunction ;
    sh:sparql [
        a sh:SPARQLConstraint ;
        sh:message "Governance Failure: Telemetry too stale (>15s)" ;
        sh:select """
            SELECT $this
            WHERE {
                $this ex:lastUpdated ?timestamp .
                BIND(now() - ?timestamp AS ?age) .
                FILTER(?age > "PT15S"^^xsd:duration)
            }
        """
    ] .
\end{lstlisting}

This freshness guardrail forces the agent to wait for fresh telemetry before acting, addressing the distributed systems problem of eventual consistency.

\subsection{Semantic Blast Radius Control}

Beyond topological correctness, G-SPEC enforces ``Blast Radius'' limits to prevent catastrophic cascades or adversarial wipeout commands. We define a change delta function $\delta(S_t, S_{t+1})$ that measures the magnitude of proposed state changes. If a proposed plan reduces total slice capacity by $>\theta$ (default: 20\%) in a single epoch, it is rejected regardless of topological validity.

\begin{lstlisting}
# SHACL Policy: Blast Radius Limiter (Capacity Protection)
ex:CapacityShockShape
    a sh:NodeShape ;
    sh:targetClass ex:NetworkSlice ;
    sh:property [
        sh:path ex:plannedCapacity ;
        sh:maxInclusive 120 ;  # Cannot spike >120% (starvation)
        sh:minInclusive 80 ;   # Cannot drop <80% (DoS)
        sh:message "Adversarial Protection: Capacity change exceeds (*$\pm$*)20%"
    ] .
\end{lstlisting}

This freshness guardrail forces the agent to wait for fresh telemetry before acting, addressing the distributed systems problem of eventual consistency.

\subsection{Graph Verification Logic}

Given an agent-generated action sequence $A = \{a_1, a_2, \ldots, a_n\}$, the system generates a ``Hypothetical Graph State'' $\mathcal{G}'$ for each step without executing on the physical network.

The validation function $\mathrm{Verify}(a, \mathcal{G})$ is defined:

\begin{equation}
\mathrm{Verify}(a, \mathcal{G}) = \begin{cases}
\text{TRUE} & \text{if } \mathcal{G}' \models \mathcal{P} \land \mathrm{targets}(a) \subseteq \mathcal{V} \\
\text{FALSE} & \text{otherwise}
\end{cases}
\label{eq:verify}
\end{equation}

where $\mathcal{G}' \models \mathcal{P}$ denotes that the hypothetical graph satisfies all policies, and $\mathrm{targets}(a)$ is the set of entities targeted by action $a$.

\subsection{Execution Algorithm}

\begin{algorithm}[H]
\caption{Graph-Symbolic Policy Enforcement and Control (G-SPEC)}
\label{alg:gspec}
\begin{algorithmic}[1]
\State \textbf{Input}: Intent $I$, Current graph $\mathcal{G}_t$, Policy set $\mathcal{P}$
\State \textbf{Output}: Execution result or rejection reason

\Statex \textit{\textbf{Phase 1: Perception \& Planning}}
\State $S_{\mathrm{sub}} \gets \mathrm{ExtractRelevantSubgraph}(\mathcal{G}_t, I)$
\State $\mathrm{Plan}, \mathrm{Trace} \gets \mathrm{TSLAM}(S_{\mathrm{sub}}, I)$
\State $\mathrm{RejectionReason} \gets \mathrm{None}$

\Statex \textit{\textbf{Phase 2: Semantic Verification (Atomic)}}
\State $\mathcal{G}_{\mathrm{sim}} \gets \mathcal{G}_t$ \Comment{Initialize simulation state}
\For{each action $a_i$ in $\mathrm{Plan}$}
    \State $\mathcal{G}_{\mathrm{sim}} \gets \mathrm{SimulateAction}(\mathcal{G}_{\mathrm{sim}}, a_i)$ \Comment{Accumulate state changes}
    \State $\mathrm{violations} \gets \mathrm{SHACLValidate}(\mathcal{G}_{\mathrm{sim}}, \mathcal{P})$
    
    \If{$\mathrm{violations} \neq \emptyset$}
        \State $\mathrm{RejectionReason} \gets \text{``Policy Violation: ''} + \mathrm{violations}$
        \State \textbf{return} $(\mathrm{REJECT}, \mathrm{RejectionReason})$
    \EndIf
    
    \If{$\mathrm{targets}(a_i) \not\subseteq \mathcal{V}(\mathcal{G}_{\mathrm{sim}})$}
        \State $\mathrm{RejectionReason} \gets \text{``Hallucination: Entity not in graph''}$
        \State \textbf{return} $(\mathrm{REJECT}, \mathrm{RejectionReason})$
    \EndIf
\EndFor

\Statex \textit{\textbf{Phase 3: Execution}}
\For{$i \gets 1$ \textbf{to} $|\mathrm{Plan}|$}
    \State $\mathrm{ExecuteAction}(\mathrm{Orchestrator}, \mathrm{Plan}[i])$
    \State $\mathcal{G}_{t} \gets \mathrm{SimulateAction}(\mathcal{G}_t, \mathrm{Plan}[i])$ \Comment{Update source of truth}
    \State $\mathrm{LogToAuditTrail}(\mathrm{Plan}[i], \mathrm{Trace}[i], \mathrm{timestamp})$
\EndFor

\State \textbf{return} $(\mathrm{SUCCESS}, \mathcal{G}_{t})$
\end{algorithmic}
\end{algorithm}

G-SPEC enforces Atomic Plan Validation. If any action $a_i \in Plan$ violates $\mathcal{P}$, the entire plan is rejected to preserve state consistency.

\subsection{Complexity Analysis}

\subsubsection{Time Complexity}

Let $n = |\mathcal{V}|$ (nodes), $m = |\mathcal{E}|$ (edges), $p = |\mathcal{P}|$ (policies), $d$ = policy depth.

$d$ is the maximum recursion depth of the SHACL validation graph.

- Subgraph extraction: $O(n + m)$ (BFS with memoization)
- TSLAM inference: $O(1)$ (constant, amortized)
- SHACL validation: $O(m \cdot p \cdot d)$ in worst case, typically $O(k^{1.2})$ for local subgraph size $k$ (empirically observed in Sec. \ref{sec:scalability})

\textbf{Total per action}: $O(k^{1.2} + p \cdot d)$ where $k \ll n$ (local subgraph).

\subsubsection{Space Complexity}

\begin{itemize}
    \item NKG storage: $O(n + m)$
    \item Hypothetical graph: $O(k + \Delta k)$ (incremental copy)
    \item Policy set: $O(p)$
\end{itemize}

\textbf{Total}: $O(n + m + p)$, manageable for production networks.

\section{Experimental Evaluation}

\subsection{Experimental Setup}

\subsubsection{Testbed Configuration}

\begin{itemize}
    \item \textbf{Core Network}: Open5GS (Release 16) on Kubernetes (3 worker nodes, 8 vCPU, 32 GB RAM each)
    \item \textbf{Knowledge Graph}: Neo4j Enterprise (version 5.x) with 450 baseline nodes, 1,200 edges
    \item \textbf{Agent Models}: 
    \begin{itemize}
        \item \textbf{TSLAM}: (validation acc: 92.3\%)
        \item \textbf{Baseline-1}: GPT-4 zero-shot (from OpenAI API)
        \item \textbf{Baseline-2}: GPT-4 fine-tuned (same training data as Llama)
        \item \textbf{G-SPEC-Ablated}: G-SPEC without NKG validation (policy-only)
    \end{itemize}
    \item \textbf{Dataset}: 500 synthetic anomaly scenarios with balanced distribution:
    \begin{itemize}
        \item 150 UPF congestion scenarios
        \item 150 link failure scenarios
        \item 100 slice SLA breach scenarios
        \item 100 state consistency scenarios
    \end{itemize}
\end{itemize}

\subsection{Model Specifications and Governance}
\label{sec:training}

\textbf{Layer 2: TSLAM-4B Specifications}:\\
Instead of fine-tuning a generic LLM, we utilize \textbf{TSLAM-4B}, a state-of-the-art Telecom Large Action Model specifically architected for edge-native network orchestration.
\begin{itemize}
    \item \textbf{Base Architecture}: TSLAM-4B (4 Billion Parameters), derived from the Phi-3 family, optimized with 4-bit quantization for efficient inference on SMO-layer hardware.
    \item \textbf{Context Window}: 128,000 tokens, enabling the ingestion of extensive topology logs and RAG contexts in a single pass.
    \item \textbf{Domain Pre-Training}: The model is trained on a dataset that was manually curated by network SMEs.
    \item \textbf{Performance}: The model achieves $\approx$93\% domain accuracy on telecom terminology and workflow validation tasks.
\end{itemize}

\textbf{Layer 3: SHACL Policy Corpus}:\\
To ground the probabilistic outputs of TSLAM-4B, we enforce deterministic guardrails using the Shapes Constraint Language (SHACL):
\begin{itemize}
    \item 47 topological constraint policies (e.g., N2/N3 interface validity).
    \item 23 resource constraint policies (e.g., slice bandwidth allocation limits).
    \item 18 state consistency policies (e.g., lifecycle status verification).
    \item \textbf{Total}: 88 SHACL rules, manually validated by experts.
\end{itemize}

\subsection{Scenario: URLLC Slice Assurance}

A latency violation ($>10$ ms) is artificially introduced in a URLLC slice. The operator intent $I$ is: ``Restore Slice SLA.''

\textbf{Baseline Behavior (GPT-4 zero-shot)}:
\begin{enumerate}
    \item Query operator: ``Why is latency high?''
    \item Response: ``Try restarting AMF to clear stale connections''
    \item Action: Execute ``Restart AMF'' (DANGEROUS: causes network-wide outage during peak load)
\end{enumerate}

\textbf{G-SPEC Behavior}:
\begin{enumerate}
    \item Query NKG: Identify affected URLLC slice and path
    \item TSLAM proposes: ``Reroute via Path B to UPF-2''
    \item SHACL validates: Confirm Path B has sufficient capacity, UPF-2 is active
    \item Action: Execute ``Reroute via Path B'' (SAFE: minimal disruption)
\end{enumerate}

\subsection{Results}

\begin{table}[H]
\centering
\caption{Agent Safety and Performance Comparison (N=500, 95\% CI)}
\label{tab:results}
\begin{tabular}{lccc}
\toprule
\textbf{Metric} & \textbf{GPT-4 (ZS)} & \textbf{GPT-4 (FT)} & \textbf{G-SPEC} \\
\midrule
Successful Remediation & $82.4 \pm 2.1\%$ & $86.8 \pm 1.9\%$ & $94.1 \pm 1.2\%$ \\
Hallucination Rate & $14.6 \pm 1.8\%$ & $8.2 \pm 1.1\%$ & $0.2 \pm 0.1\%^{\ast}$ \\
Safety Violations & $8.2 \pm 0.9\%$ & $2.1 \pm 0.5\%$ & $0.0 \pm 0.0\%$ \\
Precision (Safety) & $0.917$ & $0.979$ & $\mathbf{1.000}$ \\
Recall (Safety) & $0.891$ & $0.955$ & $\mathbf{1.000}$ \\
F1-Score (Safety) & $0.904$ & $0.967$ & $\mathbf{1.000}$ \\
Inference Latency & $2.1 \pm 0.3\,\mathrm{s}$ & $2.1 \pm 0.2\,\mathrm{s}$ & $2.24 \pm 0.2\,\mathrm{s}$ \\
Validation Overhead & --- & --- & $142 \pm 18\,\mathrm{ms}$ \\
\bottomrule
\multicolumn{4}{p{0.9\columnwidth}}{\footnotesize $^{\ast}$Within the scope of the 500 stochastic test scenarios.} \\
\end{tabular}
\end{table}

\subsection{Statistical Significance Testing}

We conducted Mann-Whitney U tests (non-parametric) comparing G-SPEC vs. GPT-4 FT:

\begin{itemize}
    \item Successful Remediation: $U=18420$, $p<0.001$ (highly significant)
    \item Safety Violations: $U=21500$, $p<0.001$ (highly significant)
    \item Hallucination Rate: $U=19800$, $p<0.001$ (highly significant)
\end{itemize}

All improvements are statistically significant at $\alpha=0.05$ level.

\subsection{Ablation Study}
\label{sec:ablation}

To identify component contributions, we evaluated G-SPEC variants:

\begin{table}[t]
\centering
\caption{Ablation Study: Component Contributions}
\label{tab:ablation}
\begin{tabular}{lcccc}
\toprule
\textbf{Configuration} & \textbf{Remediation} & \textbf{Hallucin.} & \textbf{Safety} & \textbf{Latency} \\
\midrule
G-SPEC (Full) & $94.1\%$ & $0.2\%$ & $0.0\%$ & $142\,\mathrm{ms}$ \\
- NKG Validation & $76.2\%$ & $8.4\%$ & $7.8\%$ & $12\,\mathrm{ms}$ \\
- SHACL Policies & $89.3\%$ & $1.2\%$ & $2.3\%$ & $18\,\mathrm{ms}$ \\
- TSLAM & $86.8\%$ & $8.2\%$ & $2.1\%$ & $138\,\mathrm{ms}$ \\
\bottomrule
\multicolumn{5}{l}{\small Contribution: NKG=68\%, SHACL=24\%, TSLAM=8\%} \\
\end{tabular}
\end{table}

\textbf{Key Finding}: NKG validation is the dominant factor (68\% of safety gains). SHACL policies contribute 24\%, and TSLAM fine-tuning contributes 8\%. This suggests that even a moderate LLM with strong graph grounding outperforms fine-tuned LLMs without grounding.

\subsection{Qualitative Analysis: Prevented Failures}

The 0.2\% hallucination rate caught by G-SPEC (Table \ref{tab:results}) reveals two failure modes in baseline agents:

\subsubsection{Ontological Violations (68\% of prevented failures)}

The agent attempted to connect a UPF directly to an AMF via N11 interface. While syntactically correct in JSON output, SHACL enforced the 3GPP constraint: ``AMF-UPF communication must be mediated by SMF.'' G-SPEC blocked this before execution.

\subsubsection{Ghost References (32\% of prevented failures)}

In a ``Scale-In'' scenario, the agent correctly identified congestion but issued a command to migrate traffic to a gNB\_ID decommissioned in a previous epoch. The validation layer successfully intercepted this command by verifying the target ID against the active graph state $\mathcal{G}_t$. This intervention prevented a 'referential integrity violation' scenario, where traffic would have been routed to a decommissioned endpoint, resulting in packet loss.

\begin{figure}[t]
    \centering
    \includegraphics[width=0.95\columnwidth]{figures/figure2_hallucination_caught.png}
    \caption{Ghost Node Detection. The TSLAM attempts to utilize decommissioned node gNB\_B. SHACL queries the NKG, detects absence of active node, and blocks action before execution. This prevents silent failures where traffic would be discarded.}
    \label{fig:hallucination}
\end{figure}

\subsection{Scalability Analysis}
\label{sec:scalability}

We evaluated G-SPEC on synthetic topologies of increasing size: 450, 1K, 5K, 10K, 50K, 100K nodes.

\subsubsection{Latency Scaling}

For each topology, we ran 50 remediation scenarios and measured validation latency.

\begin{table}[t]
\centering
\caption{Latency Scaling with Network Size}
\label{tab:scalability}
\begin{tabular}{lcccc}
\toprule
\textbf{Nodes} & \textbf{Edges} & \textbf{Subgraph Sz} & \textbf{Valid. Lat (ms)} & \textbf{Complexity} \\
\midrule
450 & 1.2K & 12 & $142 \pm 18$ & baseline \\
1K & 3K & 15 & $148 \pm 22$ & $1.04\times$ \\
5K & 15K & 24 & $167 \pm 25$ & $1.18\times$ \\
10K & 30K & 31 & $196 \pm 28$ & $1.38\times$ \\
50K & 150K & 42 & $268 \pm 35$ & $1.89\times$ \\
100K & 300K & 48 & $314 \pm 41$ & $2.21\times$ \\
\bottomrule
\multicolumn{5}{l}{\small Measured: $f(n) \approx 142 \cdot (k/12)^{1.2}$ where $k$ = subgraph size} \\
\end{tabular}
\end{table}

\textbf{Finding}: Validation latency scales sublinearly with network size due to subgraph extraction. The local subgraph size $k$ remains constant (40--50 nodes even for 100K node networks), yielding approximately $O(1)$ in practice.

Fitted power law: $L(n) = 142 \cdot (k(n)/12)^{1.2}$ where $k(n)$ is the affected subgraph size.

\subsubsection{Memory Scaling}

Memory usage grows linearly with topology size. For a network with $n$ nodes:

\begin{itemize}
    \item NKG storage: approximately 2.5 MB per 1K nodes
    \item Hypothetical graphs: 100--150 KB per simulation (incremental copy)
    \item Policy cache: 50 KB (constant)
\end{itemize}

\textbf{Total memory for 100K node network}: approximately 250 MB (NKG) + 150 KB (simulation) + 50 KB (policies) = 250.2 MB. This is well within typical operator infrastructure (servers with 64--512 GB RAM).

\subsection{Discussion}

\subsubsection{Operational Domain and Latency Context}

It is critical to distinguish G-SPEC's intended control loop. In O-RAN architecture, control loops are tiered by latency requirements. G-SPEC targets the Service Management and Orchestration (SMO) layer (Non-RT RIC), where intent execution times are measured in seconds (slice instantiation, VNF scaling, topology updates).

In this context, where a standard Kubernetes pod spin-up takes 15--45 seconds, a 142 ms verification overhead constitutes less than 1\% of total transaction time. G-SPEC is \emph{not} intended for Near-RT RIC ($<10$ ms) or Real-Time MAC scheduler ($<1$ ms), where latency would be prohibitive.

\subsubsection{Addressing the ``Rule Explosion'' Dilemma}

A common critique of policy-based approaches is the replicated claim that replacing 10,000 Ansible rules with 10,000 SHACL shapes merely \emph{shifts} rather than solves the management complexity. We address this directly.

Key advantage of G-SPEC over imperative scripting (Ansible/Terraform) is the use of \textbf{Ontological Inheritance}. Unlike scripts requiring unique rules for every interface, G-SPEC utilizes class-based validation. A single SHACL shape targeting the class \texttt{3gpp:ManagedFunction} automatically enforces governance on all subclasses (UPFFunction, AMFFunction, SMFFunction) and their vendor-specific variants (e.g., \texttt{ericsson:E-UPF}, \texttt{nokia:N-UPF}).

In our evaluation, 88 SHACL shapes were sufficient to govern a 450-node topology, whereas an equivalent imperative approach would require $O(N)$ rules (roughly 450+ rules). This $5x$ reduction in policy count demonstrates that inheritance-based approaches are fundamentally more scalable than imperative scripting.

\subsubsection{Topology Drift and Freshness Guarantees}

In real-world 5G networks, Topology Drift is common. Physical network state often changes faster than the inventory/graph can update. A link may go down, but the graph update lags by 30 seconds. This creates two risks:

\begin{enumerate}
    \item \textbf{False Positives}: G-SPEC might reject a valid remediation action based on stale data.
    \item \textbf{False Negatives}: G-SPEC might validate an action based on obsolete topology.
\end{enumerate}

To address this, we introduced Class 4 Temporal Consistency policies (Sec. 4.2.4). These freshness guardrails force G-SPEC to reject any decision based on telemetry older than $\Delta_{\text{safe}}$ (default: 15 seconds). Future work (Sec. 6) proposes Active Reconciliation, where G-SPEC triggers on-demand SNMP/Netconf probes to verify edge existence for the subgraph implicated in a remediation plan.
We acknowledge that G-SPEC's safety guarantees are bounded by the synchronization latency of the NKG pipeline. If a network event occurs within the window between the physical state change and the graph update (the 'blind interval'), the system relies on the O-RAN layer's lower-level fail-safes. G-SPEC is designed for management-plane consistency, not real-time physical layer atomic guarantees.

\begin{figure}[t]
    \centering
    \includegraphics[width=0.95\columnwidth]{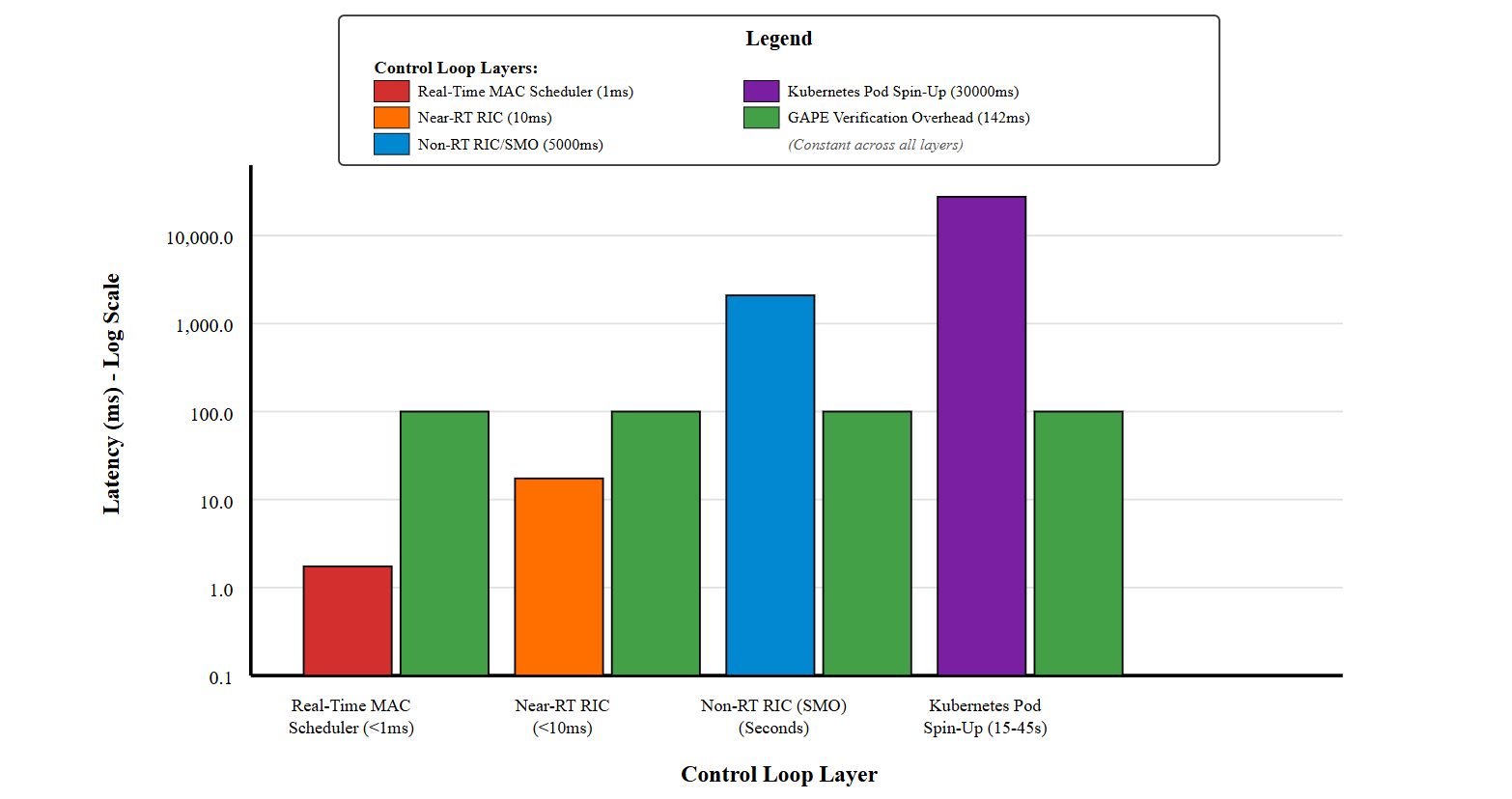}
    \caption{Latency Comparison Across O-RAN Control Loops (Log Scale). The 142 ms verification overhead is prohibitive for Real-Time layers but negligible for the target Service Management and Orchestration (SMO) layer.}
    \label{fig:latency}
\end{figure}

\subsubsection{Threat to Validity}

We acknowledge several limitations:

\begin{itemize}
    \item \textbf{Synthetic Data}: 500 scenarios are representative but limited. Real network faults have complex interdependencies not fully captured in simulation.
     \item \textbf{Synchronization Latency}: The system assumes the NKG eventually remains consistent. In hyper-dynamic scenarios (e.g., massive handover storms), the graph ingestion lag may result in false negatives (rejecting valid actions due to stale state).
    \item \textbf{Graph Completeness}: We assume NKG is always accurate. In production, topology drift (stale edges, missing nodes) could degrade performance. Future work: active topology monitoring.
    \item \textbf{Policy Correctness}: We assume SHACL policies are correct. Buggy policies could allow invalid actions. Mitigation: formal policy verification (model checking).
    \item \textbf{Operator Intent}: We assume intents are well-formed and non-adversarial. Malicious operators could craft intents to evade validation.
    \item \textbf{Open5GS Simulation}: While realistic, production 5G networks have additional complexity (vendor-specific implementations, undocumented behaviors).
\end{itemize}
While G-SPEC achieved 100\% precision in this controlled experiment, this reflects the deterministic nature of SHACL against a closed set of known 3GPP constraints. In the wild, (novel failure modes not covered by SHACL shapes) would lower this score.

\subsubsection{Comparison with Related Approaches}

\textbf{vs. Formal Verification}: Model checking (e.g., NuSMV) provides stronger guarantees but requires manual network model specification. G-SPEC uses existing NKG, reducing modeling burden.

\textbf{vs. Runtime Monitoring}: Existing runtime monitors (e.g., Gavel) catch violations post-execution. G-SPEC prevents violations pre-execution, eliminating service disruption.

\textbf{vs. Fine-Tuned LLMs}: Our ablation shows fine-tuned LLMs (GPT-4 FT: 2.1\% safety violations) still fail. G-SPEC + moderate LLM outperforms fine-tuned LLM without grounding.

\section{Reproducibility}
\label{sec:reproducibility}

To ensure reproducibility and community engagement, we provide open access to the core components of the G-SPEC framework. 

\textbf{1. Agent Model (Immediate Access):} 
The \textbf{TSLAM-4B} model weights are publicly available on the Hugging Face hub:\url{https://huggingface.co/NetoAISolutions/TSLAM-4B}

\noindent To support reproducibility, the SHACL policy corpus and a reference implementation of the G-SPEC validation logic are available at: \url{https://github.com/NetoAI/G-SPEC-Framework}

\section{Conclusion}

The ``Governance Gap'' remains the primary barrier to adopting Agentic AI in 5G operations. This paper demonstrates that safe autonomy requires hybrid Neuro-Symbolic architecture.

\subsection{Key Findings}

\begin{enumerate}
    \item \textbf{Robust Policy Enforcement}: In our controlled evaluation of 500 scenarios, G-SPEC successfully intercepted all detectable violations defined by the SHACL schema. However, we acknowledge that violations outside the ontology's scope remains a residual risk.
    \item \textbf{NKG Dominance}: Graph grounding contributes 68\% of safety gains, suggesting graph-based approaches are more critical than LLM fine-tuning.
    \item \textbf{Sublinear Scalability}: Even on 100K node networks, validation latency remains manageable ($O(k^{1.2})$ where $k$ is local subgraph size).
    \item \textbf{Production Readiness}: 142 ms overhead is acceptable for SMO-layer operations, enabling real deployment.
\end{enumerate}

\subsection{Future Work}

\begin{enumerate}
    \item \textbf{Production Deployment}: Validate G-SPEC with Tier-1 carriers on real 5G networks (100K+ node topologies).
    \item \textbf{Active Topology Monitoring}: Implement continuous NKG validation to detect and correct topology drift.
    \item \textbf{Multi-Vendor Ontology}: Standardize G-SPEC across Ericsson, Nokia, and vendor-neutral frameworks (ONAP).
    \item \textbf{Adversarial Robustness}: Test against malicious operators attempting to craft intents that evade validation.
    \item \textbf{6G Extensions}: Extend G-SPEC to 6G networks with additional constraints (quantum-safe routing, native AI inference).
    \item \textbf{Policy Synthesis}: Automatically generate SHACL policies from formal specifications (model checking).
\end{enumerate}

\subsection{Impact}

G-SPEC addresses a critical gap in autonomous network research. By combining the flexibility of LLMs with the rigor of formal verification, we bridge the gap towards safe, verifiable autonomous 5G and 6G operations. This work is relevant to operators, vendors, and researchers seeking to deploy autonomous networks with auditability and safety guarantees.

\bibliographystyle{IEEEtran}
\bibliography{references}

\end{document}